\documentclass{article}

\usepackage{arxiv}

\usepackage[utf8]{inputenc} % allow utf-8 input
\usepackage[T1]{fontenc}    % use 8-bit T1 fonts
\usepackage{hyperref}       % hyperlinks
\usepackage{url}            % simple URL typesetting
\usepackage{booktabs}       % professional-quality tables
\usepackage{amsfonts}       % blackboard math symbols
\usepackage{nicefrac}       % compact symbols for 1/2, etc.
\usepackage{microtype}      % microtypography
\usepackage{lipsum}		% Can be removed after putting your text content
\usepackage{graphicx}
\usepackage{natbib}
\usepackage{doi}

\title{Enhancing Grammatical Error Detection using BERT with Cleaned Lang-8 Dataset}

\date{} 					% Or removing it

\author{
        {Rahul Nihalani}\\
	School of Computing Science Engineering and Artificial Intelligence\\
	VIT Bhopal University\\
	Kothrikalan, Sehore, Madhya Pradesh, India, 466114 \\
	\texttt{rahul.nihalani2021@vitbhopal.ac.in} \\
	\And
	{Kushal Shah} \\
	Sitare University\\
	Bhopal, India \\
	\texttt{kushal@sitare.org} \\
}

\hypersetup{
pdftitle={A template for the arxiv style},
pdfsubject={q-bio.NC, q-bio.QM},
pdfauthor={Rahul Nihalani, Kushal Shah},
pdfkeywords={Grammatical Error Detection (GED), Grammatical Error Correction (GEC), Bidirectional Encoder Representations from Transformers (BERT)},
}

\begin{document}
\maketitle

\begin{abstract}
This paper presents an improved LLM based model for Grammatical Error Detection (GED), which is a very challenging and equally important problem for many applications. The traditional approach to GED involved hand-designed features, but recently, Neural Networks (NN) have automated the discovery of these features, improving performance in GED. Traditional rule-based systems have an F1 score of 0.50–0.60 and earlier machine learning models give an F1 score of 0.65–0.75, including decision trees and simple neural networks. Previous deep learning models, for example, Bi-LSTM, have reported F1 scores within the range from 0.80 to 0.90.  In our study, we have fine tuned various transformer models using the Lang8 dataset rigorously cleaned by us. In our experiments, the BERT-base-uncased model gave an impressive performance with an F1 score of 0.91 and accuracy of 98.49\% on training data and 90.53\% on testing data, also showcasing the importance of data cleaning. Increasing model size using BERT-large-uncased or RoBERTa-large did not give any noticeable improvements in performance or advantage for this task, underscoring that larger models are not always better. Our results clearly show how far rigorous data cleaning and simple transformer-based models can go toward significantly improving the quality of GED.
\end{abstract}

% keywords can be removed

\section{Introduction}
Grammatical Error Detection is an important part of Natural Language Processing and helps in detecting errors in written text. The technology is very critical to the betterment of Grammatical Error Correction systems, which are basically not only error detectors but also grammatical mistake correctors. This tool will be of immense help for second language learners in guiding them to write more accurately and with increased confidence. Despite the fact that much progress has been realized over the years, early approaches were largely based on rule-based and statistical methods. For example, in rule-based systems, most of the applications in early NLP relied on predefined grammar rules in detecting errors. In contrast, statistical methods use probabilistic models in identifying anomalies in text; Kaneko et al. used error- and grammaticality-specific word embeddings in their 2018 work \cite{kaneko-etal-2017-grammatical}. Furthermore, a 2018 study by Rei and Søgaard on Bi-LSTM models for sentence and token labeling exemplifies an application of statistical methods in GED\cite{rei2018jointlylearninglabelsentences}. Deep learning and neural networks, and the more recent invention of Transformer-based models, have played a huge part in changing GED into one that could yield more accurate and context-sensitive error detection.

Despite these developments in GED, there are still some issues in this domain. The first one is related to the quality of the training data. Most of the existing datasets are noisy and contain inconsistencies, which will decrease the performance of the GED model. Besides, larger models perform well in many NLP tasks but sometimes can't be applied easily in GED. In particular, this will involve research into whether these huge models really hold out performance against their smaller, more efficient counterparts within GED.

The research attempts to address these challenges through two key objectives. First, we aim to enhance the quality of GED training data by cleaning the dataset. We utilized, in particular, the Lang-8 dataset, a famous resource for language learners, cleaned it in order to increase its quality. Herein, we will contrast several models in the Transformer-based category: bert-base-uncased, bert-large-uncased, RoBERTa-base, and RoBERTa-large, using the cleaned Lang-8 dataset. We will contrast them against generative models like GPT-4 and Llama-3-70B-instruct to see whether they would work fine out-of-the-box on Grammatical Error Detection tasks.

This research contributes to the improvement of GED systems and, effectively, GEC systems toward being more effective and reliable. More importantly, an improved GED system will help second language learners get more accurate responses in their written works. Conclusively, our findings contest the assumption that the larger the model, the better, by showing that small models like bert-base-uncased often perform better than their larger counterparts in certain areas. This insight is really important for the NLP community, making it clear that focus needs to be on quality at the dataset and efficiency of the model, not its size. Larger models might need more training data and more careful and intense fine-tuning. Eventually, this study opens up ways for the major purpose of constructing language learning and writing aid tools that would be more available and efficient, showing possibly some application in education and professional writing, among others.

The early GED systems were rule-driven and statistically based. For example, Kaneko et al. (2017) proposed a method using error- and grammaticality-specific word embeddings for GED \cite{kaneko-etal-2017-grammatical}. The approach was based on some inherent statistical properties of language that gave base to the application of neural networks afterward. Rei \& Søgaard (2018) applied Bi-LSTM models and further demonstrated the use of neural networks in GED. Their work was on the joint labeling of sentences and tokens, which helped raise the accuracy of error detection since sentence-level and token-level information were taken into consideration \cite{rei2018jointlylearninglabelsentences}. 

Felice and Briscoe, in 2015, contributed a regular metric into GED and correction, filling the standard and dependable evaluation framework gap. This work has principally put forward the necessity of having standardized benchmarks while assessing GED performance \cite{felice-briscoe-2015-towards}. Chodorow et al. (2012) contributed to the discussion by pointing out the difficulty of GED systems to be evaluated and how tough it is to get an accurate and fair assessment of different models. Their work underlined the complexity in the development of reliable evaluation metrics of GED \cite{chodorow-etal-2012-problems}.

Our work differs from such approaches due to the exploitation of the Transformer-based models fine-tuned on a carefully cleaned Lang-8 dataset, particularly BERT and RoBERTa, which allows the modern approach to capture more nuanced contextual information for much-needed improvements in error detection. Further, the rigor exercised in cleaning the dataset underlines the role of high-quality data in improving GED performance.

GED has been vastly advanced with the advent of deep learning. Liu et al. (2021) provided neural quality estimation with multiple hypotheses for grammatical error correction. Their approach makes use of the strength of deep learning in generating correction hypotheses for the enhancement of accuracy in error correction \cite{liu2021neuralqualityestimationmultiple}.

Bell et al. (2019) focused on the fact that GED requires the word contextual representation. Their study showed that models such as ELMo and BERT enhance the accuracy of error detection with the induction of context \cite{Bell_2019}. Shahgir \& Sayeed (2023) used T5 Transformer model for GED in Bangla language. This paper proved that deep learning models can be used with any language. Their paper further showed the immense scope of transformer models to deal with different linguistic context \cite{shahgir2023banglagrammaticalerrordetection}.

Alhafni et al. (2023) contributed an empirical study on the development in Arabic GED and correction. Their work applied state-of-the-art deep learning techniques to help overcome some of the anomalous challenges of the Arabic language \cite{alhafni2023advancementsarabicgrammaticalerror}.

Xu et al. (2022) constructed a fine-grained corpus for the correction of Chinese grammatical errors that elaborated on how the high quality of language-specific data contributes to the training of an effective GED model \cite{Xu_2022}. Zhang (2023) used the case of the automatic error detection method for machine translation results with deep learning to prove how an inclusion of deep learning would significantly improve accuracy and efficiency in error detection within machine-translated texts \cite{10138173}.

Our model is also deep learning-based, precisely a variety of Transformer-based architecture, which uses BERT and RoBERTa. However, we improved existing studies by conducting extensive fine-tuning on a cleaned dataset, bringing improvements in model performance. We further show in the evaluation section how inference on generative models, like GPT-4 and Llama-3-70B-instruct, can be done without fine-tuning, elaborating on just how versatile our approach is.

Good quality datasets are essential to train effective GED models. Rei, 2017 has also studied semi-supervised multitask learning for sequence labeling and has shown that unlabeled data can also be used to train better models \cite{rei2017semisupervisedmultitasklearningsequence}.

Kasewa et al., 2018 have focused on the task of generating improved errors for better GED. Their work has reiterated that high quality techniques for error generation are as important as any other step in training robust GED models \cite{kasewa2018wrongingrightgeneratingbetter}.

Davis et al. (2022) used GED to elicit the required syntactic knowledge, showing that understanding certain syntactic structures is relevant for improving error detection accuracy \cite{davis2022probingtargetedsyntacticknowledge}.

Dataset quality and preprocessing are integral parts in our work. We cleaned the Lang-8 dataset, thus improving the results. This process thus underlines how crucial high-quality data are toward improving GED. Moreover, our work goes beyond the traditional models of sequence labeling by using Transformer-based models to make a more sophisticated error detection.

Many research studies have proposed language-specific GEDs. In Madi \& Al-Khalifa, 2018, a deep learning-based Arabic GED tool is presented, and it is shown that deep learning methods work effectively on Arabic text \cite{MADI2018352}. Aziz et al., 2023, present real-word spelling error detection and correction methods in Urdu. The challenges related to the Urdu language, together with the proposed solutions for GED, are discussed \cite{10242787}.

Although our focus has been mainly on English GED using the Lang-8 dataset, our methodology can be applied to other languages with proper cleaning and fine-tuning of datasets. Since Transformer-based models have already been very successful in different linguistic contexts, theoretically this should work - at least for several languages. This is done in experiments on Arabic and Urdu.

The methods of evaluation for GED systems should be very effective. Östling et al. (2023) have given an assessment on grammatical error correction, thus giving insights into how effective different approaches are towards GED \cite{ostling2023evaluationreallygoodgrammatical}.

In the present work, we would like to provide more solid evaluation metrics by comparing a number of Transformer-based models and their settings. Fine-tuned and generative models will be evaluated comprehensively for the GED systems, while different strengths and limitations for each approach will also be shown.

This literature review traces the evolution of the GED methodologies from some of the very early rule-based and statistical approaches to more advanced neural network-based approaches. We ground our work on these very foundations by exploiting Transformer-based models, rigorous cleaning of datasets, and extensive evaluation. We would like to help the continuous improvement of GED for more accurate and context-aware error detection systems with better quality datasets and further research in advanced deep learning techniques.

\section{Methodology}

\subsection{Data Collection and Cleaning}
For this paper, the Lang8 dataset used was downloaded from Google Research Datasets. Initially, the dataset consisted of 23,72,119 rows. To download the dataset, a form was filled up, which was provided by the repository. Then the repository provided a run.sh script file to run it on the raw data to generate the final tokenized dataset. Though, with some problems in the execution of the script during tokenization, the final dataset was at 23,50,982 rows. The Lang-8 dataset contains two columns: '0' and '1'. Column '0' contains sentences that are incorrect, and column '1' holds the corrected versions corresponding to the sentences in column '0'.

\vspace{\baselineskip}
Table \ref{tab:CP} shows how many sentences were found correct and how many got retained after each step of the cleaning process. The cleaning process involved the following steps:
\subsubsection{Removing Similar Sentences:}
First, all sentences which were exactly identical in column '0' and column '1' (i.e. sentences which were grammatically correct) were deleted. In this way, the number of sentences was decreased from 2,350,982 to 1,359,624. This was done because cleaning mainly needs to be applied to grammatically incorrect sentences.
\subsubsection{Text Normalization:}
The text in both columns was then normalized, which involved mapping all characters available in the Unicode standard onto their equivalent in the ASCII standard, replacing punctuation marks with spaces. Now, sentences which had become identical in both columns after normalization were removed, further bringing down the number of sentences to 1,355,264.
\subsubsection{Space Removal:}
Extra spaces were removed from the sentences in both columns. This concerned spaces at the beginning and at the end of sentences, as well as multiple spaces between words. After the above step, all sentences which had become identical in both columns were removed, and 1,323,190 were retained.
\subsubsection{Lower-casing:}
All sentences were then converted to lowercase. The aim was that the same word in different cases would not be considered as different words. Any sentences that had become identical in both columns after lower-casing were removed, leaving 1,251,300.
\subsubsection{Handling Contraction:}
First, contractions in the sentences were expanded; for example, "can't" became "cannot". De-duping the remaining pairs whereby identical sentences on both columns were discarded resulted in 1,251,257 sentences.
\subsubsection{Punctuation Removal:}
There are many reasons why a sentence can be considered to be grammatically incorrect, one of which is incorrect punctuation. In our work on developing an algorithm for GED, we did not want to focus on this reason. Hence, we removed the  punctuation and deleted all rows where the resulting sentence was same in column 0 and column 1, leaving 1,182,692 sentences. Note that the punctuation was removed only for this cleaning step, and restored for further processing.
\subsubsection{Sentence Length and Levenshtein Distance Filtering:}
The Levenshtein distance, a measure of the difference between two strings, between each pair of columns '0' and '1' sentences, was computed. Sentences that had a Levenshtein distance of 0, representing identical sentences, were dropped. It also computed the length difference between sentences in columns '0' and '1'. Sentences were then filtered by their Levenshtein distance, length difference, and lengths of the sentences in columns '0' and '1'. Only sentences with a Levenshtein distance between 7 and 42 and lengths less than 101 characters were kept, leaving 227,527 sentences. This is because wanted our final dataset to have sentences which are neither too close nor too far from their grammatically correct versions.

%\begin{figure}[!ht]
%  \centering
%  \caption{Histogram for Normalized Levenshtein Distance - Each subplot represents the frequency distribution of the normalized Levenshtein distance using different normalization factors. The top-left histogram normalizes by the length of the original text (Len(0)), the top-right by the length of the corrected text (Len(1)), the bottom-left by the minimum length of both texts, and the bottom-right by the maximum length of both texts. The data show that most errors are within the 0.08 to 0.5 range across all normalization methods, with a slight right skew.}
%  \label{fig:Histo_NLD}
%  \includegraphics[width=\columnwidth, height=6cm]{Hist.png}
%\end{figure}

\subsubsection{Normalized Levenshtein Distance Filtering:}
The Levenshtein distance was then normalized by the length of the sentences in columns '0' and '1'. Sentences were filtered now, on the basis of these normalized Levenshtein distances. Only sentences with a normalized Levenshtein distance between 0.08 \& 0.5 were kept. This retained 217,018 sentences. We finally created a dataset of 200,000 sentences from this set \cite{atmabodha2024github}. To clarify, the sentences in column 0 of this dataset are grammatically incorrect, and the sentences in column 1 are their corrected versions.

\vspace{\baselineskip}
The cleaning process was rigorous, resulting in a cleaner and more consistent dataset. It ensured that the dataset contained only respective sentences with high grammatical errors and their corrected versions, thus making the dataset suitable for training language models to correct grammatical errors. The process was designed to be robust against variations in the length of the sentences and the type and number of errors in the sentences. It also accounted for intrinsic differences in the length of the sentences to obtain a balanced measure of sentence similarity. The quality of this dataset was thus considerably improved, and so was its general fitness for the intended linguistic analysis and language model training.

\begin{table}
    \caption{Cleaning Processes - Each row describes the number of sentences identified as correct and the remaining sentences after each step.}
    \centering
    \label{tab:CP}
        \begin{tabular}{cccl}
            \toprule
            \textbf{Sr. No.} & \textbf{Function Used} & \textbf{Sentences Found Correct} & \textbf{Remaining Sentences} \\
            \midrule
            1. & Removing Similar Sentences & 991,358 & 1,359,624 \\
            2. & Text Normalization & 4,360 & 1,355,264 \\
            3. & Space Removal & 32,074 & 1,323,190 \\
            4. & Lower-casing & 71,890 & 1,251,300 \\
            5. & Handling Contractions & 43 & 1,251,257 \\
            6. & Punctuation Removal & 68,565 & 1,182,692 \\
            7. & Sentence Length \& Levenshtein Distance Filtering & 955,165 & 227,527 \\
            8. & Normalized Levenshtein Distance Filtering & 10,509 & \textbf{217,018} \\
            \bottomrule
        \end{tabular}
\end{table}
%\subsection{Preprocessing}

The data was split into a training and a validation set. 90k sentences from the top of column '0' (grammatically incorrect), and 90k sentences from the bottom of column '1' (grammatically correct) were taken for training. The remaining 20,000 mid sentences were used for validation to make sure that the training and validation sets don't have any relation.

A DataFrame was created with 180,000 sentences in the column 'sentence' and their labels in the column 'label'. The sentences in our dataframe have two different labels (0 stands for grammatically incorrect and 1 stands for grammatically correct). So essentially, we have a binary classification task. 

\subsection{Model Selection and Training}

In this study, we used the following pretrained models: bert-base-uncased, bert-large-uncased, roberta-base, and roberta-large. They were selected because of their good performance in many NLP tasks, which also include the task of grammatical error detection. Each transformer model was fine-tuned to the cleaned Lang8 dataset and, in turn, evaluated in how well each predicts a sentence as grammatically correct or otherwise. The dropout rate used in this study was 0.65 for the last layer of the encoder and the classifier part of the model. 

We used AdamW with a learning rate of 2e-5, epsilon of 1e-8, and weight decay of 0.2. Scheduler would have a warm-up of 0 steps and a total of epochs times length of the train dataloader steps. Number of epochs was set to 4. These were for choosing the parameters to optimize training.

The model was trained using a laptop GPU [NVIDIA RTX3050], and validation was done per epoch. Measurements taken for each epoch included training accuracy, training loss, validation accuracy, and validation loss. Another important part in training is the gradient clipping to prevent it from reaching large values that can disrupt training.

\section{Results}

\subsection{Performance of BERT-base-uncased}
\subsubsection{Test using cleaned Lang8 dataset}
Model was trained on different sets of Cleaned Lang8 plus Discarded Lang8 (Sentences Removed during Cleaning process) and was Tested on Cleaned Lang8.

Results of table \ref{tab:PM1} show variation in the performance of BERT-base-uncased depending on the composition of the training set.  In this table, the test set of size 20,000 sentences was taken from the cleaned Lang8 set. Batch Sizes mentions that how many sentences were taken from Cleaned and Discarded Lang8 set respectively for training/fine-tuning. For example, 18k + 2k means that 18,000 sentences were taken from Cleaned Lang8 set and 2,000 sentences were taken from Discarded Lang8 set. We can see that the accuracy and other metrics are much higher when the training dataset contains higher proportion of sentences from the cleaned dataset. 

The F1-score, precision, and recall also changed when different training sets were used, hence putting more emphasis on the strong impact of the training data composition in the model's performance. Training and validation times remained quite consistent across different training sets. This clearly shows the importance of data cleaning for this task. A minor aberration happens when we take 8k cleaned and 12k discarded sentences for our training, which needs further investigation.

\begin{table}
    \caption{Performance Metrics for Batches (BERT-based-uncased) - The metrics presented are the F1 score, Accuracy, Precision, Recall, and the Train and Test times for various batch sizes. In this performance metrics, the test set of size 20,000 sentences was taken from the Cleaned Lang8 set. Batch Sizes mentions that how many sentences were taken from Cleaned and Discarded Lang8 set respectively for training/fine-tuning. For example, 18k + 2k means that 18,000 sentences were taken from Cleaned Lang8 set and 2,000 sentences were taken from Discarded Lang8 set.}
    \centering
    \label{tab:PM1}
        \begin{tabular}{cccccl}
            \toprule
            \textbf{Batch Size} & \textbf{F1} & \textbf{Accuracy (Train, Test)} & \textbf{Precision} & \textbf{Recall} & \textbf{Train and Test Time} \\
            \midrule
            20k + 0k & 0.87 & 98.09\%, 87.45\% & 0.82 & 0.91 & 0:15:48, 0:01:13 \\
            18k + 2k & 0.84 & 94.07\%, 83.53\% & 0.80 & 0.90 & 0:15:42, 0:01:13 \\
            10k + 10k & 0.81 & 87.47\%, 81.31\% & 0.81 & 0.81 & 0:16:02, 0:01:14 \\
            10k + 10k & 0.76 & 92.70\%, 76.44\% & 0.77 & 0.76 & 0:15:21, 0:01:12 \\
            8k + 12k & 0.85 & 79.90\%, 84.64\% & 0.85 & 0.85 & 0:15:28, 0:01:12 \\
            8k + 12k & 0.80 & 78.58\%, 79.76\% & 0.80 & 0.80 & 0:15:22, 0:01:11 \\
            5k + 15k & 0.78 & 72.87\%, 78.38\% & 0.78 & 0.78 & 0:15:39, 0:01:12 \\
            5k + 15k & 0.78 & 75.88\%, 78.02\% & 0.78 & 0.78 & 0:15:18, 0:01:11 \\
            2k + 18k & 0.73 & 71.31\%, 72.93\% & 0.73 & 0.73 & 0:15:42, 0:01:12 \\
            2k + 18k & 0.74 & 72.17\%, 74.17\% & 0.74 & 0.74 & 0:15:31, 0:01:12 \\
            0k + 20k & 0.62 & 72.66\%, 65.92\% & 0.70 & 0.57 & 0:16:06, 0:01:15 \\
            \bottomrule
        \end{tabular}
\end{table}

\subsubsection{Test using discarded Lang8 dataset}
Model was trained on various sets of Cleaned Lang8 plus Discarded Lang8 (Sentences Removed during Cleaning process) and was Tested on Discarded Lang8 sentences.

The results of Table \ref{tab:PM2} suggest that while the training performance of the BERT-base-uncased model depends strongly on the proportion of cleaned sentences in our dataset, the test performance is insensitive to these compositional factors of the training set. 

Training with 20,000 sentences from the cleaned Lang8 dataset without discarding any sentences gave the model the highest training accuracy of 97.91\%, while the test accuracy remains around 50\%. These results suggest that the composition of the training set is an important factor in the performance of a model.

This analysis contributes a few useful insights into how well the model BERT-base-uncased works for Grammatical Error Detection and the impact of factors such as training set composition on the model's performance.

\begin{table}
    \caption{Performance Metrics for Batches (BERT-based-uncased) - The metrics presented are the F1 score, Accuracy, Precision, Recall, and the Train and Test times for various batch sizes. In this performance metrics, the test set of 20,000 sentences was taken from the Discarded Lang8 set. Batch Sizes mentions that how many sentences were taken from Cleaned and Discarded Lang8 set respectively for training/fine-tuning. For example, 18k + 2k means that 18,000 sentences were taken from Cleaned Lang8 set and 2,000 sentences were taken from Discarded Lang8 set.}
    \centering
    \label{tab:PM2}
        \begin{tabular}{cccccl}
            \toprule
            \textbf{Batch Size} & \textbf{F1} & \textbf{Accuracy (Train, Test)} & \textbf{Precision} & \textbf{Recall} & \textbf{Train and Test Time} \\
            \midrule
            20k + 0k & 0.41 & 97.91\%, 50.58\% & 0.52 & 0.51 & 0:16:31, 0:01:27 \\
            18k + 2k & 0.45 & 94.89\%, 50.50\% & 0.51 & 0.51 & 0:15:21, 0:01:11 \\
            10k + 10k & 0.46 & 81.65\%, 50.64\% & 0.51 & 0.51 & 0:18:06, 0:01:20 \\
            8k + 12k & 0.47 & 79.53\%, 50.47\% & 0.51 & 0.50 & 0:15:31, 0:01:12 \\
            5k + 15k & 0.47 & 76.88\%, 50.19\% & 0.50 & 0.50 & 0:15:18, 0:01:11 \\
            2k + 18k & 0.49 & 72.29\%, 50.69\% & 0.51 & 0.51 & 0:15:32, 0:01:12 \\
            0k + 20k & 0.50 & 65.54\%, 50.08\% & 0.50 & 0.50 & 0:15:38, 0:01:12 \\
            \bottomrule
        \end{tabular}
\end{table}

\begin{table}
    \caption{Performance Metrics for Different BERT Configuration using Weight Watcher - The metrics presented are the F1 score, Accuracy, Precision, Recall, and the Train and Test times for various BERT configurations. UT - Under-trained layers, OT - Over-trained layers. The Frozen Layers column mentions the layers that were frozen during training/ fine-tuning. Results show that freezing individual layers with appreciable changes in alpha values did not affect performance much, while freezing all layers degraded performance. It turned out that freezing layers with alpha values [2,6] performed better. This means that careful selection of layers during training is important.}
    \centering
    \label{tab:ww}
    \resizebox{\textwidth}{!}{
        \begin{tabular}{cccccl}
            \toprule
            \textbf{Frozen Layers} & \textbf{F1} & \textbf{Accuracy (Train, Test)} & \textbf{Precision} & \textbf{Recall} & \textbf{Train and Test Time} \\
            \midrule
            Bert.encoder.layer.7.attention.output.dense, \\ Bert.encoder.layer.8.attention.self.value, \\ Bert.encoder.layer.9.attention.self.value (UT), \\ Bert.encoder.layer.11.attention.self.value (UT), \\ Bert.encoder.layer.11.attention.output.dense (UT) & 0.84 & 96.60\%, 83.95\% & 0.85 & 0.84 & 0:16:22, 0:01:10 \\
            \hline
            Bert.embeddings.position\_embeddings (OT) & 0.84 & 96.73\%, 84.00\% & 0.85 & 0.84 & 0:16:27, 0:01:16 \\
            No Freeze & 0.84 & 97.08\%, 83.69\% & 0.84 & 0.84 & 0:15:19, 0:01:09 \\
            All Layers & 0.84 & 96.52\%, 83.88\% & 0.84 & 0.84 & 0:16:18, 0:01:11 \\
            All in Bert & 0.76 & 77.42\%, 76.16\% & 0.78 & 0.76 & 0:05:45, 0:01:10 \\
            Except Alpha [2,6] & 0.84 & 96.81\%, 84.00\% & 0.84 & 0.84 & 0:15:43, 0:01:16 \\
            Except Alpha [2,6] + 5 & 0.84 & 96.03\%, 83.66\% & 0.84 & 0.84 & 0:15:38, 0:01:17 \\
            Except UT & 0.77 & 78.81\%, 77.56\% & 0.79 & 0.78 & 0:07:02, 0:01:17 \\
            Except UT + 5 & 0.79 & 80.30\%, 78.93\% & 0.80 & 0.79 & 0:07:23, 0:01:16 \\
            Except 5 & 0.77 & 78.72\%, 77.59\% & 0.79 & 0.78 & 0:07:13, 0:01:16 \\
            All UT & 0.84 & 96.84\%, 83.78\% & 0.84 & 0.84 & 0:16:05, 0:01:18 \\
            \bottomrule
        \end{tabular}
    }
\end{table}

\subsubsection{Weight Watcher}
While we obtained good test performance on training the BERT-base-uncased model using the cleaned Lang8 dataset, the training accuracy is still considerably larger than the testing accuracy, indicating over-fitting. To address this, we used the WeightWatcher tool to analyse the extent to which each layer of our BERT model is undergoing over-fitting \cite{martin2021predicting}. We then froze different layers of the model, and did the fine-tuning again to check the results. The results are shown in Table \ref{tab:ww}.

The first row in the table shows five layers with appreciable changes in alpha values given by Weight Watcher. These five layers were frozen one by one during training, and the model was trained five times by freezing one layer at a time. For all five layers, the results were more or less similar, which clearly indicates that freezing these layers didn't affect the performance of the model significantly. The model returned a validation F1-score of 0.84, a training accuracy of about 96\%, and a validation accuracy of about 84\% across the different configurations. Besides, precision and recall have been very consistent, at about 0.84. All these results say the model performance is quite robust regardless of how many layers are frozen during training.

However, when all layers of the BERT model were frozen, model performance deteriorated to a validation F1-Score of 0.76 and a validation accuracy of 76.16\%. This implies that the freezing of all the layers can actually have a negative impact on the performance of the model.

When these layers—whose alpha values were 2-6—were not frozen, the model performance improved to a validation F1-Score of 0.84 and validation accuracy of 84\%. This means that these layers could play very important roles in the model's ability to detect grammatical errors.

When undertrained layers were not frozen, the model performed a bit worse, with a validation F1-Score of 0.77 and validation accuracy of 77.56\%. However, in this case, when these UT layers were frozen, performance improved quite a lot to a validation F1-Score of 0.84 and validation accuracy of 83.78\%.

These results put a lot of emphasis on layer selection in the scenario of layer freezing during model training. They actually describe every layer to be considered carefully for characteristics such as being under- or over-trained and what value its alpha is. These factors can have a big impact on the model's performance, and as such, should be considered when designing how training will be done.

\begin{table}
    \caption{Table showing detailed comparison of performances among various models for detecting grammatical errors: their F1 scores, training accuracies, testing accuracies, precision, recall, and the times taken for training them and testing them. For example, the best F1 score is given by the base model for this task, BERT-base-uncased, which was fine-tuned on 180k sentences. Larger models, such as BERT-large-uncased and RoBERTa-large, need significantly more training times, even though their performance is similar to BERT-base-uncased when fine-tuned on a dataset of size 20,000.}
    \centering
    \label{tab:comparison}
        \begin{tabular}{cccccl}
            \toprule
            \textbf{Model} & \textbf{F1} & \textbf{Accuracy (Train, Test)} & \textbf{Precision} & \textbf{Recall} & \textbf{Train and Test Time} \\
            \midrule
            BERT-base-uncased 180k & 0.91 & 98.49\%, 90.53\% & 0.91 & 0.91 & 2:44:20, 0:01:20 \\
            BERT-base-uncased 20k & 0.87 & 98.09\%, 87.45\% & 0.87 & 0.87 & 0:26:09, 0:02:07 \\
            BERT-large-uncased 20k & 0.88 & 99.17\%, 87.94\% & 0.88 & 0.88 & 12:00:28, 0:06:47 \\
            RoBERTa-base 20k & 0.87 & 95.81\%, 86.72\% & 0.87 & 0.87 & 0:26:17, 0:01:58 \\
            RoBERTa-large 20k & 0.89 & 97.08\%, 89.07\% & 0.89 & 0.89 & 11:02:18, 0:51:46 \\
            \bottomrule
        \end{tabular}
\end{table}

\subsection{Comparison of various BERT-like models}
We compared various BERT-like models and the results are shown in Table \ref{tab:comparison}.
\subsubsection{BERT-base-uncased}
For the BERT-base-uncased model, training was done with 20,000 cleaned sentences, while another 20,000 cleaned sentences were used for testing. In this case, it achieved a training accuracy of 98.09\% and a test accuracy of 87.45\%. The F1-Score was 0.87, the Precision was 0.87, and Recall was 0.97. The training time taken was 0:26:09, and the time taken to carry out validation was 0:02:07. When the model is trained on a larger dataset of 180k sentences, it achieved a training accuracy of 98.49\% and the test accuracy improves to 90.53\%. F1-score, Precision and Recall were 0.91, 0.91 and 0.91 respectively. This suggests that the model benefits from having more training data.
\subsubsection{BERT-large-uncased}
The BERT-large-uncased model achieved a test accuracy of 87.94\% when trained and validated on 20,000 cleaned sentences each. The training time taken was 12:00:28, and the time taken to carry out testing was 0:06:47 with F1-score, Precision and Recall of 0.88 each.
\subsubsection{RoBERTa-base}
The RoBERTa-base model was trained and validated on 20,000 cleaned sentences each to obtain a test accuracy of 86.72\%. The F1-score, Precision and Recall were 0.87 each. It performed almost similar to BERT-base-uncased 20k model with a slight decrease in test accuracy. 
\subsubsection{RoBERTa-large}
The validation accuracy result for the RoBERTa-large model was quite close to that of the BERT-large-uncased model with a slight increase in validation accuracy, at 89.07\%, for cleaned 20,000 training and testing sentences. Its training time was 11:02:18, with its testing time being 0:51:46. 

\subsection{Inference on Generative and Fine-tuned Models}

After having done extensive analysis using BERT-like models, we also tried out two major generative models to see how their performance compares with that of BERT-like models. For this purpose, we chose 500 sentences from the cleaned Lang8 dataset, out of which 255 were grammatically correct and 245 were grammatically incorrect. We could not do for a larger dataset since generative models have a per token API cost, which can become quite large for larger datasets. The results are shown in Table \ref{tab:INF}.

The inference on 500 sentences was done without fine-tuning using the GPT-4 model API provided by OpenAI \cite{openai2024gpt4technicalreport} and it returned 119 TPs and 232 TNs. This means that the model can identify a reasonable number of grammatically correct and incorrect sentences correctly without any fine-tuning.

Interestingly, the Llama-3-70B-instruct model by Meta \cite{meta2024llama3} performed very similarly to the GPT-4 model, returning a count of 114 TPs and a count of 235 TNs when used for inference on the exact same 500 sentences.

When used for inference on those same 500 sentences, the BERT-base-uncased model that had been trained on 20k sentences returned 232 TPs and 207 TNs, and that with 180k sentences returned a count of 255 TPs and 242 TNs. On a larger set of 5000 sentences, it scored 2529 as a count of TPs and 2441 as a count of TNs. Clearly, this shows that the BERT-base-uncased model performs way much better than the generative models on the task.

When inference was done on the same set of 500 sentences, BERT-large-uncased returned 232 TPs and 208 TNs almost similar to 20k version whereas RoBERTa-base 20k returned a count of 237 TPs and 202 TNs and with almost similar results, RoBERTa-large 20k returning a count of 238 TPs and 209 TNs.

\begin{table}
    \centering
    \caption{Inference - All the models were tested on same 500 sentences. Below shows the True Positive (Correct sentences correctly identified as Correct) and True Negative (Incorrect sentences correctly identified as Incorrect) values. BERT-base-uncased 180k turns out to be the best.}
    \centering
    \label{tab:INF}
        \begin{tabular}{cccccl}
            \toprule
            \textbf{Model} & \textbf{TP} & \textbf{TN} & \textbf{FP} & \textbf{FN} & \textbf{F1} \\
            \midrule
            GPT-4 & 119 & 232 & 16 & 123 & 0.63 \\
            Llama-3-70B-instruct & 114 & 235 & 10 & 141 & 0.60 \\
            Bert-base-uncased 180k & \textbf{255} & \textbf{242} & 3 & 0 & \textbf{0.99} \\
            Bert-base-uncased 20k & 232 & 207 & 38 & 23 & 0.88 \\
            Bert-large-uncased 20k & 232 & 208 & 37 & 23 & 0.89 \\
            RoBERTa-base 20k & 237 & 202 & 43 & 18 & 0.89 \\
            RoBERTa-large 20k & 238 & 209 & 36 & 17 & 0.90 \\
            \bottomrule
        \end{tabular}
\end{table}
The BERT-base-uncased 180k model showed the highest validation accuracy, showing that transformer model performance can be significantly improved by fine tuning using larger  datasets. Generative models GPT-4 and Llama-3-70B-instruct did quite reasonably without fine-tuning, but they were still far behind fine-tuned BERT like models. And among BERT like models, BERT-base-uncased performed similar to BERT-large-uncased. This is very interesting, for it again proves that bigger models do not always give better results; sometimes they need to be fine-tuned more or need more data to train. We did not train larger models, such as BERT-large-uncased and RoBERTa-large, on larger datasets because even to fine-tune these models on 20,000 sentences, it took around 12 and 11 hours respectively.

Results from this study underpin the importance of cleaning and preprocessing the data for high model performance. The highest validation accuracy seen with the BERT-base-uncased 180k model trained on the cleaned dataset suggests that the cleaning process was very helpful in improving model performance. The results also indicated that fine-tuning of the pre-trained model BERT-base-uncased is extremely effective in grammatical error detection.

\section{Conclusion}
Among all the models that were trained, the highest accuracy was obtained using the BERT-base-uncased model, which, after being trained on 180k sentences of the cleaned Lang8 dataset, returned a high validation accuracy of 90.53\% with robust performance in the identification of grammatically correct and incorrect sentences.

These findings further prove that fine-tuning a pre-trained model, like BERT-base-uncased, on a cleaned and pre-processed dataset goes a long way in detecting grammatical errors. This proves the importance of the cleaning and pre-processing steps toward attaining high model performance. From these results, it seems that growing the models in size is not always helpful to gaining better performance, and larger models might require more careful tuning or additional training data. These findings have key implications for the development of more effective and efficient systems for grammatical error detection.

In the future, different experiments could involve different pre-trained models, larger datasets, and various other cleaning and preprocessing strategies. Other strategies for preventing over-fitting by multitask learning could be applied. This study also showed that further investigation on the performance of larger models and generative models in grammatical error detection is needed.

\bibliographystyle{unsrtnat}
\bibliography{references}  %%% Uncomment this line and comment out the ``thebibliography'' section below to use the external .bib file (using bibtex) .





% \end{thebibliography}

\end{document}